\documentclass[10pt,twocolumn,letterpaper]{article}

\usepackage{cvpr}              

\usepackage{graphicx}
\usepackage{amsmath}
\usepackage{amssymb}
\usepackage{booktabs}
\usepackage{multirow}
\usepackage{stfloats}

\usepackage{silence}
\makeatletter
\robustify\@latex@warning@no@line
\makeatother
\usepackage{authblk}

\usepackage[pagebackref,breaklinks,colorlinks]{hyperref}

\usepackage[capitalize]{cleveref}
\crefname{section}{Sec.}{Secs.}
\Crefname{section}{Section}{Sections}
\Crefname{table}{Table}{Tables}
\crefname{table}{Tab.}{Tabs.}

\def\cvprPaperID{*****} 
\def\confName{CVPR}
\def\confYear{2022}

\begin{document}

\title{Combining Efficient and Precise Sign Language Recognition: \\ Good pose estimation library is all you need}

\author[1,2]{Matyáš Boháček}
\author[3,4]{Zhuo Cao}
\author[1]{Marek Hrúz}


\affil[1]{
University of West Bohemia, Pilsen, Czech Republic 
}

\affil[2]{
Gymnasium of Johannes Kepler, Prague, Czech Republic
}

\affil[3]{
KU Leuven, Leuven, Belgium
}

\affil[4]{
ML6, Ghent, Belgium
}

\affil[ ]{}

\affil[ ]{\textit{The authors can be contacted at } \texttt{matyas.bohacek@matsworld.io}.}

\renewcommand\Authands{ and }

\maketitle

\begin{abstract}
\textbf{Notice: This extended abstract was presented at the CVPR 2022 AVA workshop\footnote{\url{https://accessibility-cv.github.io/}} in New Orleans, USA.}

Sign language recognition could significantly improve the user experience for d/Deaf people with the general consumer technology we use daily, such as IoT devices or videoconferencing. However, current sign language recognition architectures are usually computationally heavy and require robust GPU-equipped hardware to run in real-time. Some models aim for lower-end devices (such as smartphones) by minimizing their size and complexity, which leads to worse accuracy. This highly scrutinizes accurate in-the-wild applications. We build upon the SPOTER architecture, which belongs to the latter group of light methods, as it came close to the performance of large models employed for this task. By substituting its original third-party pose estimation module with the MediaPipe library, we achieve an overall state-of-the-art result on the WLASL100 dataset. Significantly, our method beats previous larger architectures while still being twice as computationally efficient and almost $11$ times faster on inference when compared to a relevant benchmark. To demonstrate our method's combined efficiency and precision, we built an online demo that enables users to translate sign lemmas of American sign language in their browsers. This is the first publicly available online application demonstrating this task to the best of our knowledge.

\end{abstract}

\section{Introduction}
\label{sec:intro}

Sign languages (SLs) are the primary means of communication for the d/Deaf communities. They are a form of natural language systems based on manual articulations and non-manual components. They utilize a significantly more variable and complex modality despite enabling one to convey identical semantics as the spoken and written language. With over 70 million people considering one of the approximately $300$ SLs as their native language, computational methods that would cross the bridge between written or spoken languages and SLs have been subjects of extensive study in the literature since the 1990s. Two prevalent topics concerning SLs have emerged: SL synthesis and SL recognition (SLR). Regardless of the time that has passed, these tasks are far from being solved.

In this work, we address SLR, whose objective is to translate videos of performed signs from a known set into a written form. It can be divided into isolated SLR, where only single lemmas are translated, and continuous SLR, translating unconstrained signing utterances. We attend the first of these streams: isolated SLR.

We identified that a critical problem of current light-weight SLR architectures aimed for applications in the wild on standard consumer devices (e.g., smartphones) is that they perform markedly worse compared to their heavier counterparts. We hence focus on boosting their accuracy without adding more computational demand. For this purpose, we build upon the SPOTER architecture~\cite{Bohacek_2022_WACV}, which came close to current heavy architectures' performance at a notably smaller size and computational requirements. \textit{Bohacek} \etal use a third-party pose estimation library in their architecture to represent the videos at the input with sequences of skeletal joint coordinates, as opposed to raw images with larger dimensionality like the heavy models. As there were no comparative experiments of various pose extraction toolkits, we substituted the original one and observed the change in performance. This extended abstract reports our observations up to now, but our work is still in progress.

\begin{figure}[t!]
\begin{center}
\includegraphics[width=0.8\linewidth]{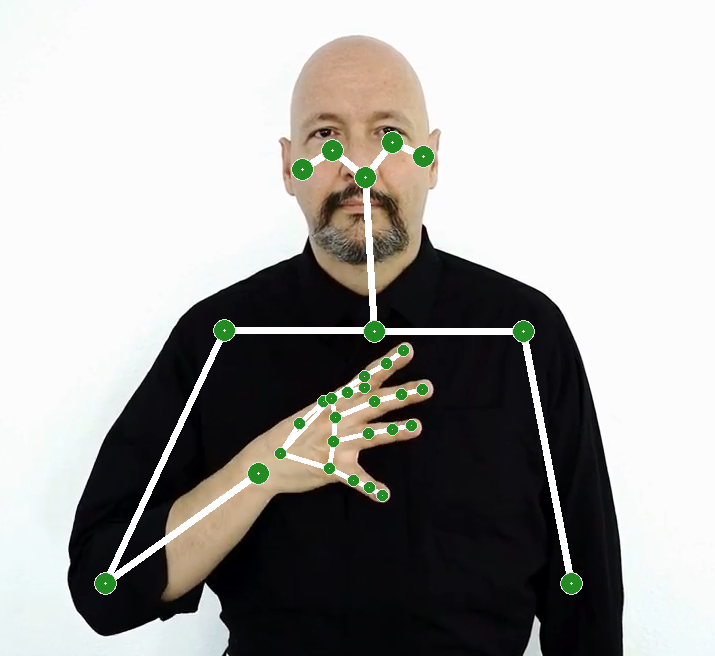}
\end{center}
\caption{Visualization of the body and hand landmarks extracted using MediaPipe and converted to the SPOTER format.}
\label{fig:pose}
\end{figure}

\begin{table*}[bp]
\begin{center}
\begin{tabular}{l|l|l}
\textbf{Model}                   & \textbf{Input type} & \textbf{WLASL100} \\ \hline
I3D (baseline) \cite{li2020word} & \multirow{3}{*}{Appearence-based}                    & 65.89             \\
TK-3D ConvNet \cite{Li2020}      &                     & 77.55    \\
Fusion-3 \cite{Hosain2021}       &                     & 75.67             \\ \hline
GCN-BERT \cite{Tunga2021}        & \multirow{4}{*}{Pose-based}                    & 60.15             \\
Pose-TGCN \cite{li2020word}      &                     & 55.43             \\
Pose-GRU \cite{li2020word}       &                    & 46.51             \\
SPOTER with Vision API (original) \cite{Bohacek_2022_WACV}                    &            & 63.18 \\ \hline
SPOTER with MediaPipe (ours)                    & Pose-based            & \textbf{78.29}

\end{tabular}
\end{center}
\caption{Top-1 macro average recognition accuracy achieved by each model on the WLASL100 subset's testing split.}
\label{tab:model_comparison}
\end{table*}

Overall, the so-far contributions of our work include:

\begin{itemize}
    \item Showing the difference that can be made by simply swapping the pose estimation library in a pose-based SLR architecture.
    \item Establishing state-of-the-art results on the WLASL100 dataset with a substantially lighter architecture than the so-far best model.
    \item Creating and open-sourcing an online demo of the model, which enables anyone to have lemmas known to our model recognized right in their browser.
\end{itemize}

\begin{figure*}[h]
  \centering
  \hfill
  \subfloat{\includegraphics[width=0.45\textwidth]{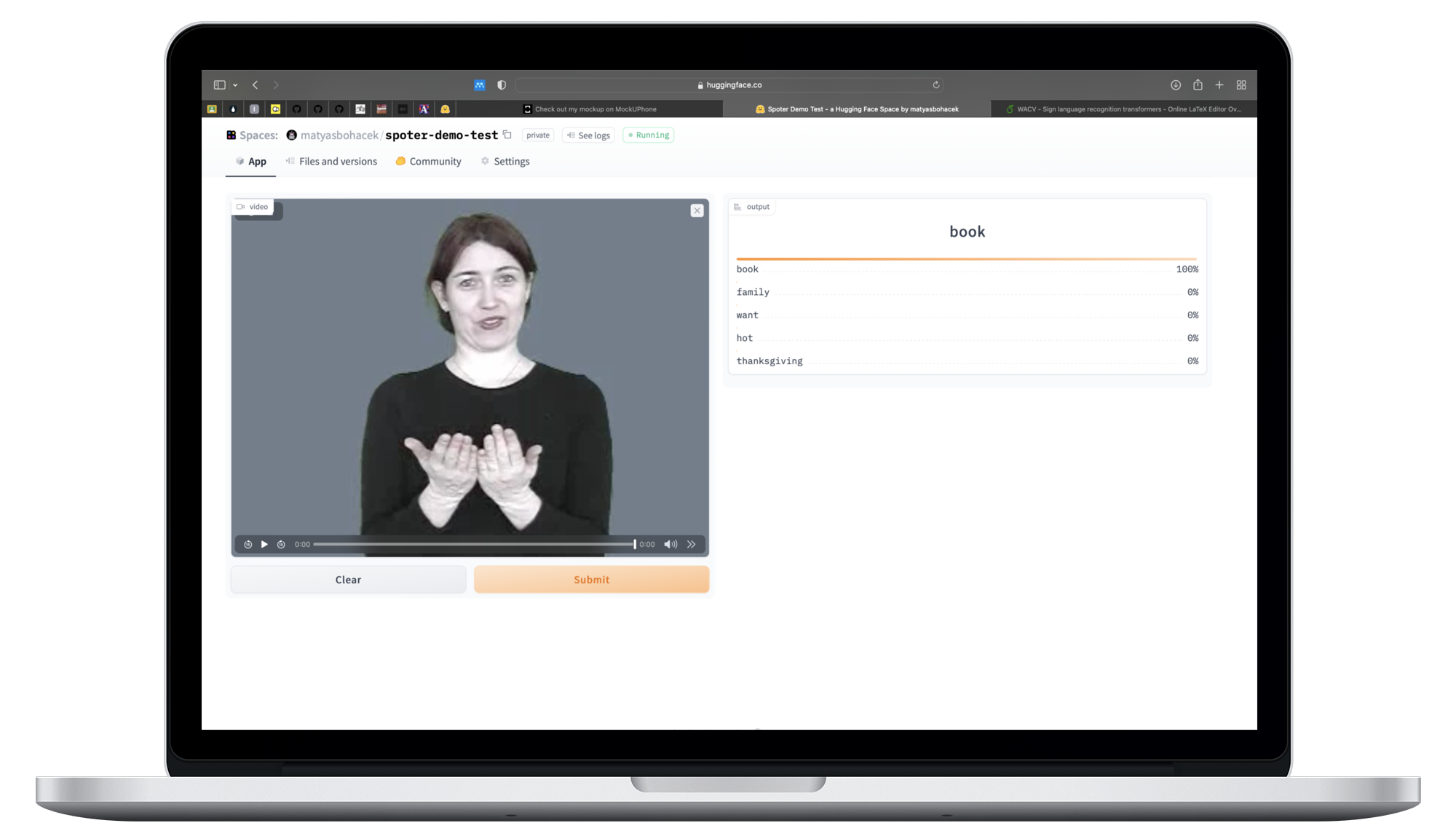}}
  \hfill
  \subfloat{\includegraphics[width=0.45\textwidth]{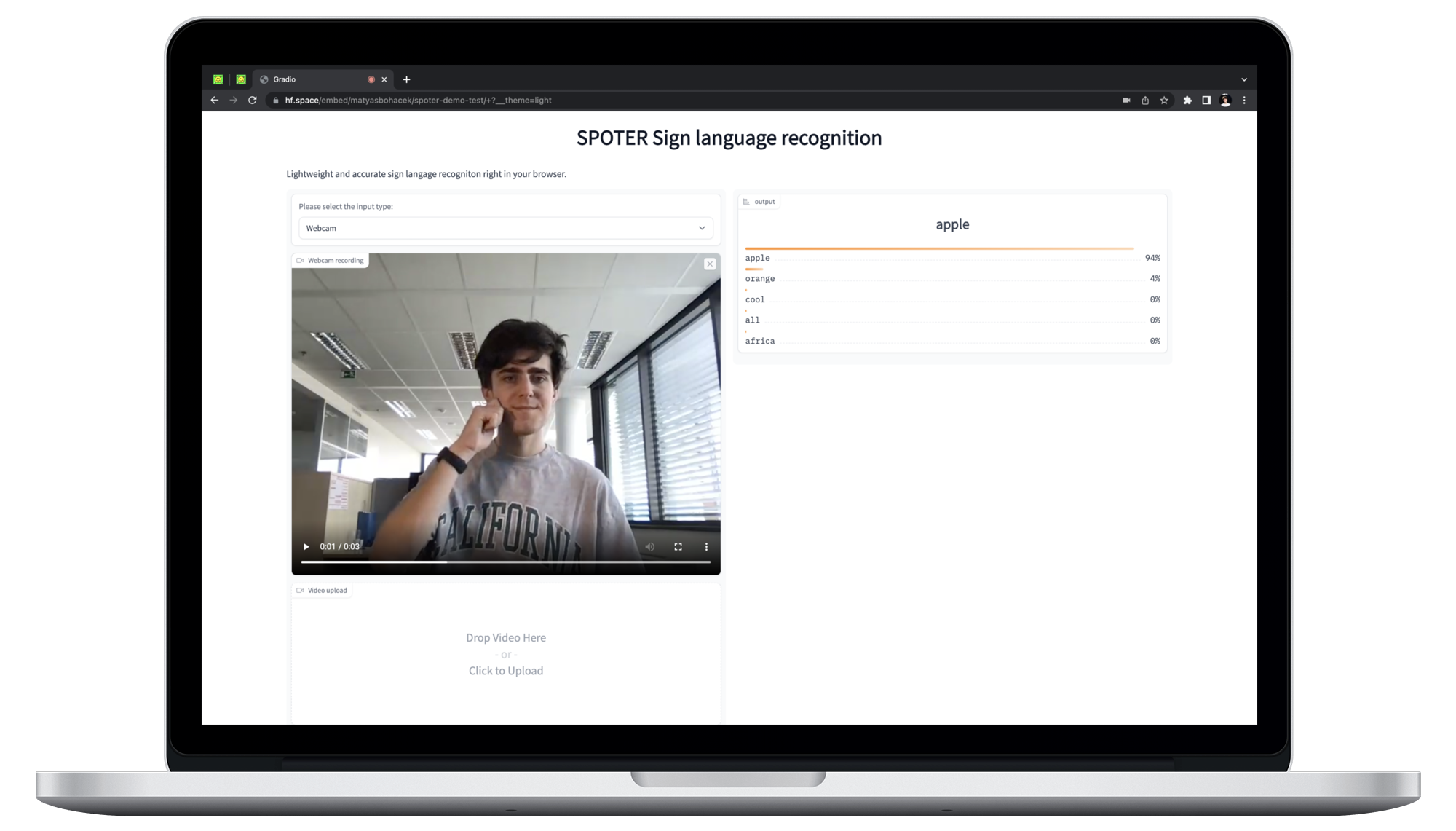}}
  \hfill
  \caption{Screenshots of the demo web application created using Gradio and hosted on Hugging Face. For a given video, the poses for each frame are extracted and propagated on the model. The user interface presents top-5 classes with percentages corresponding to the final softmax results of the model. The user can submit a pre-recorded video or record a new one directly in the browser using their webcam.}
  \label{fig:demo}
\end{figure*}

\section{Related work}
\label{sec:related-work}

The original approaches in SLR utilized shallow statistical modeling (e.g., Hidden Markov Models) that achieved reasonable performance only on small datasets containing no more than 20 classes~\cite{Starner_1997_MBR, Starner_1998_TPAMI}. More robust models operating on top of larger datasets (spanning hundreds of classes) have emerged after the dawn of deep learning. First, Convolutional neural networks (CNNs) were heavily exploited for this task~\cite{Camgoz_2020_CVPR, Pigou_2016_LREC, Saunders_2021_IJCV, Koller_2016_BMCV, Rao_2018_SPACES} by generating unified representations of the input frames that could be thereafter used for recognition. Recurrent Neural Networks (RNNs)~\cite{Cui_2017_CVPR, Koller_2020_PAMI} and Transformer-based architectures~\cite{Saunders_2021_IJCV, Camgoz_2020_ECCV, Bohacek_2022_WACV} have later been employed for this use, too. 3D CNNs~\cite{Carreira_2017_CVPR, Joze_2019_BMVC, li2020word} and Graph Convolutional Networks (GCNs)~\cite{amorim2019spatial} have been the latest architectures to be studied. These were  found to perform the best on multiple benchmarks.

Over the recent years, two general streams of works emerged in the literature differing in input representation. First is the appearance-based stream, where models expect sequences of frames as raw images (RGB or RGB-D data). These models generally perform better but are larger and computationally demanding simultaneously. Models from the second, pose-based stream, on the other hand, employ pose estimation on the input video and further analyze just the sequences of skeletal data. These models tend to be more lightweight than appearance-based ones due to the reduced dimensionality of the input data. However, they also perform notably worse.

\section{Data}
\label{sec:data}

We evaluate our model on the WLASL dataset~\cite{li2020word}, which holds over $21,000$ video instances spanning $2,000$ lemma classes from the American sign language. The videos in the dataset have been collected and manually categorized from multiple online resources, primarily dictionaries. The signers are native SL users. The dataset includes four subsets: WLASL100, WLASL300, WLASL1000, and WLASL2000, each spanning the respective number of classes. We use the first subset only. We follow the training, validation, and testing splits defined by the authors.

\section{Methods}
\label{sec:methods}

We build upon the SPOTER architecture, as proposed by \textit{Bohacek} \etal~\cite{Bohacek_2022_WACV}. We follow the implementation and run configuration from the original paper to the full extent, apart from the two changes described below.

First, we substitute the pose estimation library used in the paper (Vision API\footnote{\url{https://developer.apple.com/documentation/vision}}) with the Blazepose~\cite{bazarevsky2020blazepose} model from the MediaPipe library~\cite{lugaresi2019mediapipe}.  Both models' human pose representations (skeletal models) differ, with MediaPipe overall recognizing more landmarks but lacking the neck coordinates. As our goal is to compare the merits of the individual libraries and not those of adding new landmarks, we pre-process the MediaPipe representations to match the original structure. To do that, we only keep the landmarks present in the original SPOTER and compute the neck coordinates as the middle of the line segment between the shoulder joints. This way, we arrive at $54$ body landmarks, including $5$ head landmarks and $21$ landmarks per hand, the same as in the original architecture. The joints are depicted in Figure~\ref{fig:pose}.

Unlike in the original paper, we also employ hyperparameter search over the augmentations parameters using the Weights and Biases library~\cite{wandb}. We refer the reader to the original manuscript for a detailed description of the architecture, normalization procedure, and augmentations.

\section{Results}
\label{sec:results}

We report the accuracy of our model on the WLASL100 subset in Table~\ref{tab:model_comparison}, provided with a comparison to the previous architectures and the original SPOTER. Our SPOTER with MediaPipe achieves a testing accuracy of $78.29 \%$, constituting an overall state of the art on this benchmark. As the original SPOTER utilizing the Vision API for pose estimation achieved a testing accuracy of $63.18 \%$, we observe a boost of over $15 \%$ in absolute precision. Importantly, no modification was conducted to the architecture itself so that we can consider this purely an effect of the pose estimation library. Such a significant improvement suggests that MediaPipe evinces more accurate poses, but a detailed quantitative analysis must be conducted before conclusive inferences.

Apart from surpassing its precursor version and other pose-based methods, SPOTER with MediaPipe also outperforms all the appearance-based approaches. Even so, it is still substantially less computationally demanding. When compared to one of the appearance-based models\footnote{SPOTER was compared to the I3D model \textit{Bohacek} \etal~\cite{Bohacek_2022_WACV}.}, SPOTER has only half the number of parameters, and its inference takes on average $11$x less time.

\section{Demo}
\label{sec:demo}

To showcase that our model is highly accurate and computationally effective, we built a demo web application using Gradio~\cite{abid2019gradio}, whose screenshots are presented in Figure~\ref{fig:demo}. Therein, users can insert a pre-recorded video file of a person signing or record a new one using their webcam. Once submitted, the users are presented with the top-5 lemma classes and their probabilities from the model's final softmax layer, as predicted by SPOTER.

We believe that this application on its own can already serve as a great educational resource for those learning SLs. With slight modifications, it could be used as a search tool for online SL dictionaries or as a web interface when a closed set of responses from the user is expected, for example. Most importantly, it signals that SLR technology can work in the browser at reasonable prediction accuracy.

The demo is hosted on Hugging Face Spaces~\cite{wolf2019huggingface} and publicly available at \url{https://demo.signlanguagerecognition.com}.

\section*{Conclusion}
\label{sec:conclusion}

In this work-in-progress paper, we experimented with substituting the pose estimation library in a prominent pose-based SLR architecture. We show that by swapping Vision API with MediaPipe, SPOTER achieves an overall state-of-the-art testing accuracy of $78.29 \%$ on WLASL100 dataset. We built the first publicly available SLR demo in the browser to demonstrate the joint computational efficiency and precision of our approach. Overall, we believe this is a crucial step toward highly accurate models that are efficient enough to run on consumer devices and help break down technological barriers for d/Deaf users.

We plan to continue working on this problem by evaluating all major pose estimation libraries in a similar fashion. We want to conduct more thorough experiments and qualitative analyses, too.

{\small
\bibliographystyle{ieee_fullname}
\bibliography{egbib}
}

\end{document}